\documentclass[conference]{IEEEtran}
\IEEEoverridecommandlockouts
\usepackage{cite}
\usepackage{comment}
\usepackage{booktabs}
\usepackage{multirow}
\usepackage{soul}
\usepackage{hyperref}
\usepackage{amsmath,amssymb,amsfonts}
\usepackage{algorithmic}
\usepackage{graphicx}
\usepackage{textcomp}

\usepackage{xcolor}
\usepackage{eso-pic}
\def\BibTeX{{\rm B\kern-.05em{\sc i\kern-.025em b}\kern-.08em
    T\kern-.1667em\lower.7ex\hbox{E}\kern-.125emX}}

\definecolor{verylightred}{RGB}{255,210,220}

\newcommand{\acceptnote}{%
  \AddToShipoutPictureBG*{%
    \AtPageUpperLeft{%
      \raisebox{-1.5\baselineskip}{%
        \makebox[\paperwidth]{\footnotesize\itshape
          Accepted at the 2026 IEEE International Conference on Data Mining (ICDM 2026)}}}}}

\begin{document}
\acceptnote

\title{Beyond Tokenization: Direct Timestep Embedding and Contrastive Alignment for Time-Series Question Answering\\

}

\author{
\IEEEauthorblockN{Yafeng Wu\textsuperscript{*}, Huu Hiep Nguyen, Thin Nguyen, Hung Le}
\IEEEauthorblockA{Deakin Applied Artificial Intelligence Initiative, Deakin University, Geelong, Australia}
\thanks{\textsuperscript{*}Corresponding author: Yafeng Wu (email: s225635478@deakin.edu.au)}
}

\maketitle

\begin{abstract}
 Recent advances in large language models (LLMs) have given rise to time-series question answering (TSQA), which formulates time-series analysis as natural-language question answering. However, directly feeding raw numerical series into LLMs suffers from a tokenization bottleneck: Byte Pair Encoding fragments continuous values into unstable tokens whose embeddings lack meaningful metric structure, resulting in the loss of magnitude, scale, and trend information. Prior methods use patch-based encoders that split the series into fixed windows, locking in one granularity that breaks patterns and hides exact timesteps, through a separate module that rarely transfers across datasets with different lengths or sampling rates. To address this challenge, we propose CADE (Contrastive Alignment with Direct Embedding), a novel framework for TSQA built upon two key components: direct timestep embedding and semantic alignment. The proposed framework maps each timestep directly into the LLM embedding space through a point-wise linear encoder and MLP projector, preserving exact index-level access while eliminating the need for patching and padding. To further bridge the semantic gap between time-series and language representations, we introduce a novel one-directional supervised contrastive loss that aligns time-series embeddings with frozen class-name text anchors.
Experimental results on the public Time-MQA benchmark demonstrate that our framework consistently improves performance across six TSQA tasks, outperforming both open-source and proprietary LLM baselines.
\end{abstract}
\section{Introduction}

\begin{figure}
    \centerline{\includegraphics[width=1.05\columnwidth]{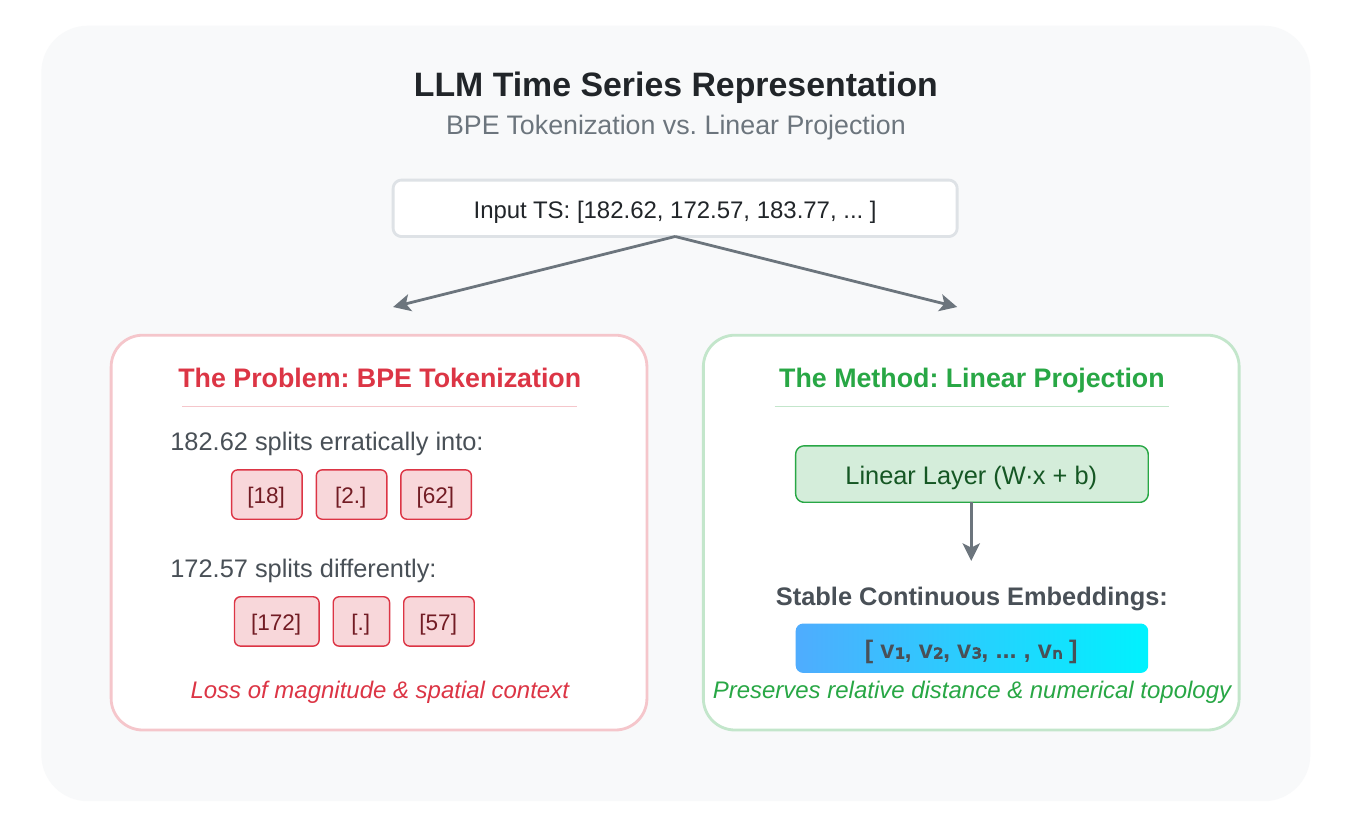}}
    \caption{Comparison of time series representation strategies.}
    \label{fig:introduction}
\end{figure}

Time series analysis underpins critical tasks such as anomaly detection, classification, forecasting, and imputation, with applications across healthcare monitoring, industrial maintenance, and financial modeling~\cite{wen2023transformers}. Traditional deep learning models have achieved strong results in capturing temporal dependencies~\cite{zeng2023transformers,wu2023timesnet,nie2023a}. Meanwhile, Large Language Models (LLMs) have shown remarkable reasoning and generalization across diverse domains~\cite{openai2023gpt4, deepseek2024v3, yang2025qwen3}. This has motivated efforts to extend their capabilities to time series. Most existing approaches use LLMs as feature extractors or auxiliary modules rather than end-to-end solvers~\cite{jin2024time, sun2024test, zhou2023one}. A more ambitious paradigm is time series question answering (TSQA). It recasts each task as a natural-language question and lets the LLM produce the answer directly. For example, a 3D gait time-series classification task can be framed as a question, with the LLM determining whether the sample indicates freezing of gait. Feeding raw time series into an LLM raises a fundamental representational problem: standard tokenizers struggle to encode continuous numerical values.

The core difficulty lies in how LLMs represent continuous numerical values, and it originates at the level of tokenization. Byte Pair Encoding (BPE) \cite{sennrich2016neural} is the standard tokenizer in most LLMs. It builds its vocabulary from frequency patterns in text, where numbers are sparse and long-tailed. As a result, it merges digits according to textual co-occurrence rather than place value: a value such as 182.62 may be split into fragments like [18], [2.], [62]. This segmentation is also context-dependent, meaning that the same digits may be tokenized differently across contexts. Thus, a single value is fractured into unstable symbols that carry no metric structure, preventing the model from receiving a stable signal that two numbers are close in magnitude. Consequently, the model struggles to recover magnitude, scale, or trend. These properties are essential for time-series reasoning, which limits the reliability of current methods on TSQA.

To bypass tokenization, prior LLM-based time series methods~\cite{wang2025itformer,jin2024time,Xie_2025} adopt patch-based encoders. These encoders segment the series into fixed-length windows and project each window into a continuous embedding. This commits the model to a single temporal granularity fixed before training. The patch length may impose boundaries that cut across meaningful patterns and fold distinct timesteps into one token. As a result, the model cannot natively identify the exact temporal index. This patch-based encoder is also a separate, often pretrained module \cite{wang2025itformer}. It adds parameters and introduces a domain-specific inductive bias that may not transfer across datasets with different lengths or sampling rates.

We instead propose CADE (Contrastive Alignment with Direct Embedding), which uses a linear projection to map each time-series value into a continuous embedding, illustrated in Figure~\ref{fig:introduction}. Unlike BPE, which fractures a value into unstable sub-tokens, our direct mapping keeps each value intact as a single continuous vector. It also places numerically close values close in the embedding space, thereby restoring the metric structure discarded by text tokenization. Each token corresponds one-to-one with a timestep, allowing the model to address individual indices directly. This is essential for tasks such as imputation that operate on specific positions. Furthermore, our approach requires neither a predefined patch size nor padding. Since the token count equals the sequence length, the same projection can process series of arbitrary lengths and sampling rates without re-segmentation or per-dataset re-tuning. It also preserves the original temporal resolution and allows the LLM's attention to model both local and long-range dependencies directly.

To further align time-series embeddings with the LLM's embedding space, we add an auxiliary one-directional supervised contrastive loss constructed from time-series classification data. For each classification sample, the loss pulls the time-series embedding toward its class-name text embedding and away from other classes. The text anchors remain frozen, so only the time-series representations are aligned with the LLM's lexical space. Although this signal comes solely from classification samples, the encoder and projector are shared by all six tasks. Therefore, the loss regularizes the shared pathway rather than fitting classification alone. This produces more discriminative and semantically grounded representations.

The main contributions of this work are as follows:
\begin{itemize}
    \item To the best of our knowledge, this is among the first works to use a lightweight linear encoder, rather than a patch-based encoder, to map time-series values into continuous embeddings for multi-task time-series question answering. Despite its simplicity, this design substantially improves LLM performance across TSQA tasks.

    \item We introduce a novel one-directional supervised contrastive loss that aligns projected time-series embeddings with frozen class-text anchors, strengthening the semantic correspondence between time-series features and the LLM's linguistic reasoning.

    \item Through extensive experiments on the Time-MQA datasets, we show that CADE achieves competitive or superior performance against both open-source and proprietary LLMs. The source code for CADE and all relevant baseline models are openly accessible on https://github.com/YafengWu/CADE.
\end{itemize}
\section{Related Work}

\subsection{Large Language Models for Time Series Analysis}

Recent work has explored adapting LLMs for time series tasks through two main directions. First, prompt-based methods such as PromptCast~\cite{xue2023promptcast} and LLMTime~\cite{gruver2023large} serialize numerical sequences into text prompts for direct forecasting. Second, multimodal architectures like Time-LLM~\cite{jin2024time} and UniTime~\cite{liu2024unitime} align temporal representations with language model embeddings for cross-domain forecasting. A parallel line of work develops dedicated time-series foundation models such as TimesFM~\cite{das2024decoder}, Chronos~\cite{ansarichronos}, and Time-MOE~\cite{shi2025time} pre-train transformers specifically on temporal data for the forecasting task. These models are primarily designed for forecasting rather than text-based question answering, so they cannot be used for question answering and are not compared in our work. Among the remaining two directions, multimodal methods do not fully leverage the reasoning and generation capabilities of pretrained LLMs, while prompt-based methods suffer from imprecise numerical representation. This motivates a unified approach that enables LLMs to directly reason over time series inputs and solve diverse temporal tasks.

\subsection{Temporal Question Answering}
Several recent works study question answering over time series data.
ChatTS \cite{Xie_2025} is a time-series multimodal LLM trained on synthetic time-series–text pairs for temporal understanding and reasoning, targeting open-ended time-series reasoning rather than the task-oriented, multi-task setting we study.
ChatTime \cite{wang2025chattime} is a pretrained time-series foundation model for forecasting and time-series QA.
We focus on adapting general-purpose LLMs to time-series question answering and therefore compare only against existing LLM-based adaptation methods rather than time-series foundation models.
Time-MQA \cite{kong2025time} introduces the Time-MQA dataset and uses LoRA to directly fine-tune an LLM across a diverse set of temporal tasks, such as classification, anomaly detection, and forecasting.
ITFormer \cite{wang2025itformer} adapts the Q-Former structure for the Time-Series Question Answering task and releases the EngineMT-QA dataset.

Despite their progress, these methods inherit the representational limitations discussed above: Time-MQA feeds numerical values through the LLM's text tokenizer and thus suffers from unstable digit segmentation and the absence of number-line geometry, while ITFormer and ChatTS rely on patch-based encoders that fix a single temporal granularity and obstruct per-timestep generation. In contrast, we map each timestep directly into the LLM's embedding space with a lightweight linear encoder and MLP projector, avoiding both tokenization artifacts and patch-level granularity constraints.

Beyond representation, aligning time series and natural language, inspired by vision–language alignment in models such as LLaVA~\cite{liu2023visual} and InstructBLIP~\cite{dai2023instructblip}, is itself an active direction \cite{liu2025towards}, with approaches based on cross-attention~\cite{jin2024time,liu2024time}, contrastive learning~\cite{chen2025ts,dong2025teaching,sun2024test}, and knowledge distillation~\cite{liu2025efficient}. However, existing contrastive methods such as TS-CLIP~\cite{chen2025ts}, TimesCLIP~\cite{dong2025teaching}, and TEST~\cite{sun2024test} either apply contrastive learning to narrow tasks (e.g., classification or forecasting) without an LLM, or treat the LLM merely as a pattern extractor by appending task-specific heads. In contrast, we align timestep embeddings with the LLM's lexical space using a one-directional supervised contrastive loss. This objective can be viewed as class-prototype alignment and is applied across six tasks within a unified framework.

\section{Method}

\subsection{Problem Formulation}
We address \emph{time-series question answering}, in which a LLM is asked to answer natural-language questions grounded in a
univariate time series. Each sample is a triple $(\mathbf{x}, q, \tau)$,
where $\mathbf{x} = \{x_1, x_2, \dots, x_T\}$ with $x_t \in \mathbb{R}$ is a
univariate time series of length $T$, $q$ is a natural-language prompt that
poses a question about $\mathbf{x}$ and contains statistical features, and
$\tau$ denotes the task type. The series length $T$ varies across samples.

We consider six task types,
$\tau \in \{$anomaly detection, classification, multiple choice,
true/false, forecasting, imputation$\}$, and unify them under a single
generative interface: regardless of $\tau$, the model produces a textual response $R = \{r_1, \dots, r_L\}$ autoregressively, rather than relying on task-specific output heads, where $L$ denotes the number of tokens in the generated answer. Formally, the model defines a
conditional distribution
\begin{equation}
p_\theta(R \mid \mathbf{x}, q)
= \prod_{i=1}^{L} p_\theta\!\left(r_i \mid r_{<i}, \mathbf{x}, q\right),
\label{eq:autoregressive}
\end{equation}
where $\theta$ is the parameter of the LLM. The objective is to maximize the likelihood of the textual response
$R$ given the time series and prompt. 

\subsection{Preprocess}
\textbf{Normalization. } $z$-score normalization is used for each series independently within its own question. Specifically, for a time series
$\mathbf{x} = \{x_1, \dots, x_T\}$, each value is normalized using the mean
and standard deviation computed from that same series:
\begin{equation}
    x'_t = \frac{x_t - \mu}{\sigma},
\end{equation}
where $x_t$ is the original value at time step $t$, $x'_t$ is the normalized
value, the full normalized series is denoted
$\mathbf{x}' = \{x'_1, \dots, x'_T\}$, and $\mu$ and $\sigma$ are
respectively, the mean and standard deviation of the time series $\mathbf{x}$.

\textbf{Statistical Features.} Instance-level $z$-score normalization removes absolute scale and offset
information that can be critical for some tasks. For example, multiple choice and true/false question tasks have some questions about statistical features such as mean and maximum values. To preserve this information, we augment the text prompt with
statistical summaries computed from the \emph{raw} (pre-normalization)
series. After the normalized series is presented, we append the following
to the textual part of the prompt:
\begin{quote}
\itshape
The above is the normalized time series data. Its raw data has the following
statistical information:\\
mean: \texttt{<mean\_val>}, standard deviation: \texttt{<std\_val>},
minimum: \texttt{<min\_val>}, maximum: \texttt{<max\_val>},
median: \texttt{<median\_val>}.
\end{quote}
This design lets the model reason over the shape of the normalized signal
while still having access, in textual form, to the absolute statistics that
normalization discards.

\subsection{Architecture}
\begin{figure*}
    \centerline{\includegraphics[scale=0.89]{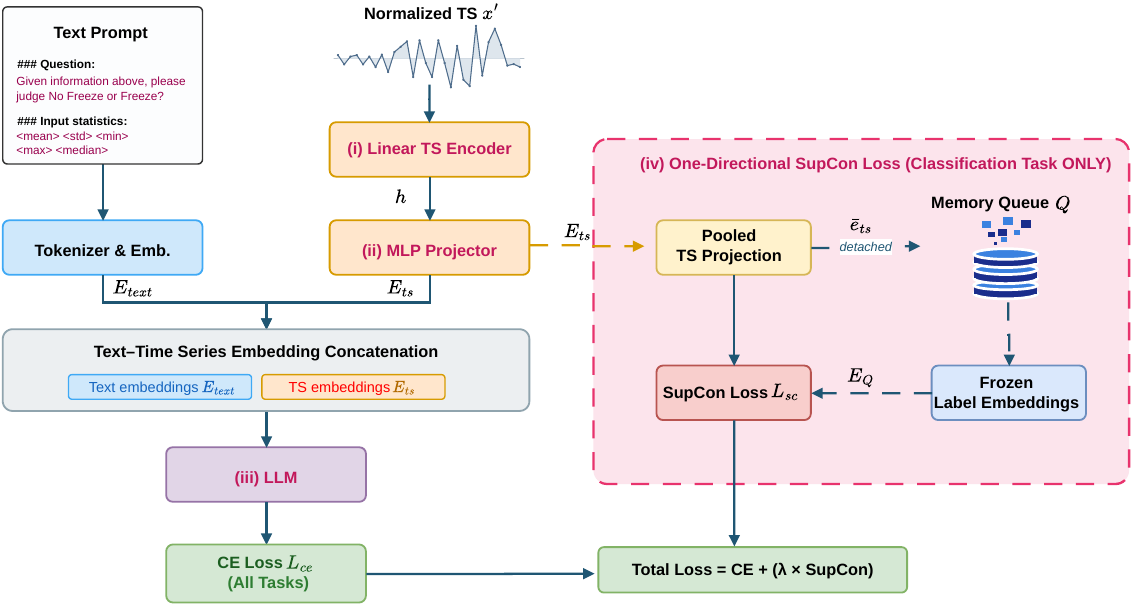}}
    \caption{Framework of the proposed CADE.}
    \label{fig:model}
\end{figure*}

As illustrated in Figure~\ref{fig:model}, the proposed framework CADE consists of four
components: (i) a \emph{linear time-series encoder} that maps the normalized
series into a continuous feature space; (ii) an \emph{MLP projector} that bridges
the time-series and language modalities by projecting the encoded features into
the LLM embedding space; (iii) an \emph{LLM} module: a pretrained LLM that autoregressively generates the answer from the fused time-series and text tokens; and (iv) a
\emph{one-directional SupCon loss} that aligns the projected
time-series embeddings with frozen class-text embeddings for the classification
task. We describe the first three components below, and detail the contrastive
module in Section~\ref{sec:od_supcon}.

\textbf{Linear TS Encoder. }
Following the design philosophy of vision--language models such as
LLaVA~\cite{liu2023visual}, which bridge a non-text modality into an LLM
through an image encoder followed by a projection module, CADE maps
the time series into the language model with an analogous
\emph{encoder--projector} pair: a linear time-series encoder produces a
continuous feature sequence, and an MLP projector
aligns it with the LLM embedding space. Concretely, a single linear layer
maps the normalized time series $\mathbf{x}' \in \mathbb{R}^{T \times 1}$
into a $d_{ts}$-dimensional continuous feature space, yielding a sequence
of encoded features $h \in \mathbb{R}^{T \times d_{ts}}$:
\begin{equation}
     h = \mathrm{Linear}(\mathbf{x}'), \qquad h \in \mathbb{R}^{T \times d_{ts}}.
\end{equation}

This linear TS encoder has three advantages. It is \emph{index-aligned}: the one-to-one correspondence between embeddings and
timesteps preserves exact positional addressability, which the generation
tasks (forecasting and imputation) exploit. It is \emph{resolution-agnostic}: by adopting point-wise tokenization,
the encoder preserves the completeness of temporal information \cite{shi2025time} and handles
series of arbitrary length and sampling rate without patching, padding, or
per-dataset re-tuning, which enhances the model's flexibility and broad
applicability to variable-length sequences while keeping the LLM's attention
operating at the signal's native resolution. Finally, it is \emph{lightweight}: a single linear layer
introduces far fewer parameters than a patch-based transformer encoder and
requires no separate pretraining, yet, as our experiments show, suffices to
inject the time series into the LLM effectively.

\textbf{MLP Projector. }
The encoded features are then projected into the LLM embedding space with an
MLP, mapping each time step embedding from $d_{ts}$ to $d_{llm}$ dimensions:
\begin{equation}
     E_{ts} = \mathrm{MLP}(h), \qquad E_{ts} \in \mathbb{R}^{T \times d_{llm}},
\end{equation}
where $d_{llm}$ denotes the token-embedding dimension of the backbone LLM, so that each of the $T$ time-series tokens is mapped into the LLM's lexical space and can be concatenated with the text embeddings. The MLP has the following structure:
\begin{align}
    \mathbf{z} &= \mathrm{GELU}\!\big(\mathbf{W}_1\, h + \mathbf{b}_1\big),
        & \mathbf{W}_1 &\in \mathbb{R}^{4\,d_{ts} \times d_{ts}}, \\
    E_{ts}     &= \mathrm{LayerNorm}\!\big(\mathbf{W}_2\, \mathbf{z} + \mathbf{b}_2\big),
        & \mathbf{W}_2 &\in \mathbb{R}^{d_{llm} \times 4\,d_{ts}}.
\end{align}
a non-linear projection provides the capacity to bridge the
two modalities. The final LayerNorm stabilizes training.

\textbf{LLM. }
A pretrained LLM generates answers from the fused time-series and text tokens:
\begin{equation}
    R = p_{\theta}\big(f(E_{ts}, E_{text})\big),
\end{equation}
where $R$ is the generated response, $p_{\theta}$ is the LLM
decoder with parameters $\theta$, defined in formula \ref{eq:autoregressive}, $E_{ts}$ is the time-series embedding, and
$E_{text}$ is the text embedding. The sequence fusion $f(\cdot)$ depends on the task.
For time-series understanding tasks (classification, anomaly detection,
multiple choice, true/false), we use $E_{ts}$ to replace the raw time series. Raw numerical values are difficult for LLMs to interpret directly; substituting the learned embedding provides a more language-compatible representation that the model can reason over more effectively.
For time-series generation tasks (forecasting and imputation), we send
$E_{ts}$ and the raw time series together: the TS embedding provides
high-level temporal understanding, while the raw values give the LLM
the precise numerical grounding it needs to perform the task.

\subsection{One-Directional Supervised Contrastive Loss}
\label{sec:od_supcon}

To better regularize the projected time-series embeddings toward the semantic structure of the LLM's embedding space, we introduce a one-directional supervised contrastive loss. Among the six tasks, only classification possesses a finite set of discrete, mutually exclusive class labels, making it uniquely suited for this objective. We therefore apply the loss exclusively to classification samples: it pulls each sample's projected time-series embedding toward its corresponding class-text embedding and pushes it away from those of other classes. We call this loss \emph{one-directional} because
gradients flow only from the time-series side toward the frozen text targets,
and never into the LLM's embedding layer. The one-directional supervised contrastive Loss
consists of four components, described below.

\textbf{Pooled TS Projection.}
During the forward pass, each sample is checked for whether it belongs to the
classification task; non-classification samples are skipped entirely and
contribute no contrastive signal. For each classification sample, we obtain
the contrastive anchor by mean-pooling the projected time-series sequence
$E_{ts} = \{\mathbf{e}_{ts}^{(1)}, \dots, \mathbf{e}_{ts}^{(T)}\}$ over
the time dimension:
\begin{equation}
    \bar{\mathbf{e}}_{ts}
    = \frac{1}{T}
      \sum_{t=1}^{T}
      \mathbf{e}_{ts}^{(t)},
\end{equation}
where $\mathbf{e}_{ts}^{(t)} \in \mathbb{R}^{d_{\text{llm}}}$ is the $t$-th
output of the MLP projector. Each pooled TS projection is paired with its class label
$c \in \mathcal{C}$ for use in the subsequent steps.

\textbf{Memory Queue.}
In a multi-task training regime with six tasks, any given mini-batch may
contain few or even zero classification samples, leaving insufficient
anchors to form meaningful positive--negative contrasts. To address this, we
maintain a circular memory queue
$Q = \bigl\{(\bar{\mathbf{e}}_{ts}^{(k)},\; c^{(k)})\bigr\}_{k=1}^{K}$
of capacity $M$, where each entry stores a \emph{detached} pooled TS projection
$\bar{\mathbf{e}}_{ts}^{(k)}$ paired with its class label $c^{(k)}$. This detachment prevents gradients from propagating through historical samples stored in the queue. Here $K \le M$ denotes the current occupancy, which grows until the queue reaches capacity $M$ and is thereafter held at $M$ by evicting the oldest entry in FIFO order.
A warmup threshold $M_{\min}$ is enforced: if $|Q| < M_{\min}$, the current
batch's anchors are enqueued but the contrastive loss is not computed,
ensuring a sufficiently diverse pool of negatives before the signal is
applied. Following standard MoCo practice~\cite{he2020momentum}, entries are
enqueued \emph{after} loss computation so that a sample is never used as its
own key.

\textbf{Frozen Label Embeddings.}
At training time, all class labels stored in the memory queue are mapped to
their corresponding frozen label embeddings.
Specifically, for each class $c \in \mathcal{C}$, a single label embedding
$\mathbf{e}_c \in \mathbb{R}^{d_{\text{llm}}}$ is constructed once before
training by tokenizing the lowercased label text and averaging the
corresponding input-embedding vectors:

\begin{equation}
    \mathbf{e}_c
    = \frac{1}{|t_c|}
      \sum_{j=1}^{|t_c|}
      \mathrm{Embed}\!\bigl(t_c^{\,j}\bigr),
\end{equation}
where $t_c^{\,j}$ is the $j$-th token of class name $c$ and
$\mathrm{Embed}(\cdot)$ denotes the LLM input-embedding layer.

The set of label embeddings is then defined as
\begin{equation}
    E_Q = \{\mathbf{e}_{c^{(k)}}\}_{k=1}^{K}.
\end{equation}
Here, $E_Q$ denotes the set of label embeddings corresponding to the class labels stored in the memory queue.
Each $\mathbf{e}_c$ is detached from the computation graph and receives no
gradient updates throughout training. In contrast to the queue detachment, this detachment keeps the label embeddings as fixed semantic anchors during training. This is precisely what makes the loss
\emph{one-directional}: the time-series projections learn to align with the
text targets, but the text targets themselves remain fixed anchors in the
shared embedding space.

\textbf{SupCon Loss.}
Given the current pooled TS projection $\bar{\mathbf{e}}_{ts}$ and the
pool of frozen label embeddings $E_Q$ retrieved from the queue, both sides
are $\ell_2$-normalized and cosine similarities are scaled by a temperature
$\gamma$:
\begin{equation}
    s^{(k)}
    = \frac{\bar{\mathbf{e}}_{ts} \cdot \mathbf{e}_{c^{(k)}}}
           {\bigl\|\bar{\mathbf{e}}_{ts}\bigr\|\;
            \bigl\|\mathbf{e}_{c^{(k)}}\bigr\|\;\gamma},
\end{equation}
yielding a similarity vector $\mathbf{s} \in \mathbb{R}^{K}$.
A positive mask is then constructed: entry $k$ is marked positive if the
queue entry $k$ shares the same class label as the anchor. We then adopt the supervised
contrastive (SupCon) formulation~\cite{khosla2020supervised}, which treats
\emph{all} queue entries sharing the anchor's class label as positives:
\begin{equation}
    \mathcal{L}_{\mathrm{sc}}
    = -\frac{1}{|\mathcal{P}|}
       \sum_{k \in \mathcal{P}}
       \log
       \frac{\exp(s^{(k)})}
            {\displaystyle\sum_{j=1}^{K}\exp(s^{(j)})},
\end{equation}
where $\mathcal{P} = \{k : c^{(k)} = c\}$ is the set of queue entries
sharing the same class as the anchor, and the denominator sums over
\emph{all} queue entries.

\subsection{Training Loss}

\textbf{Cross-Entropy Loss. }
All six tasks are trained with the standard next-token cross-entropy loss.
Given the model response $R$ and the ground-truth answer $A$:
\begin{equation}
    \mathcal{L}_{ce} = \mathrm{CrossEntropy}(R,\, A).
\end{equation}

\textbf{Total Loss. }
The final training objective combines the generative and contrastive terms:
\begin{equation}
    \mathcal{L} = \mathcal{L}_{ce} + \lambda_\tau \cdot \mathcal{L}_{sc},
\end{equation}
where
\begin{equation}
    \lambda_\tau =
    \begin{cases}
        \lambda & \text{if } \tau = \text{classification},\\
        0       & \text{otherwise},
    \end{cases}
\end{equation}
Here $\lambda \in [0,1]$ is a scalar weight that controls the strength of the contrastive term. The contrastive branch introduces no additional trainable parameters; it reuses the encoder and projector from the main forward path. Gradients from $\mathcal{L}_{sc}$ therefore update only the TS Encoder and MLP Projector, while $\mathcal{L}_{ce}$ updates the encoder, projector, and the LLM jointly.

\section{Experiments}
\subsection{Dataset}

We evaluate our method on the Time-MQA Dataset \cite{kong2025time}, a multi-task question-answering benchmark for time series understanding that covers six tasks: classification, anomaly detection (AD), true/false, multiple choice (MCQ), forecasting, and imputation. The training and testing sets are randomly sampled, with the training set containing approximately 1,400 samples per task (ranging from 1,293 for forecasting to 1,400 for classification, anomaly detection, and imputation), totaling 8,286 samples, and the test set containing approximately 400 samples per task (ranging from 379 for forecasting to 400 for classification, anomaly detection, and true/false), totaling 2,376 samples.

\subsection{Baselines}

We compare against four external methods that represent distinct design choices for time-series question answering. Our own internal variants, which degrade individual components of CADE, are introduced separately in the architectural ablations (Section~\ref{sec:arch_abl}).

\textbf{Time-MQA} \cite{kong2025time}: an LLM fine-tuned with LoRA, jointly trained on all six tasks as a unified multi-task model.

\begin{table}
\centering
\caption{Implementation details.}
\label{tab:impl_details}
\begin{tabular}{ll}
\toprule
\textbf{Configuration} & \textbf{Value} \\
\midrule
Framework & PyTorch \\
GPU & NVIDIA A100 (40GB) $\times$ 1 \\
LLM & Qwen-3-0.6B \\
Batch size & 32 \\
Learning rate & $5 \times 10^{-5}$ \\
LR scheduler & Cosine \\
Training steps & 2,000 \\
TS Encoder dimension $d_{ts}$ & 384 \\
Memory bank size & 512 \\
Contrastive loss weight ($\lambda$) & 0.1 \\
\bottomrule
\end{tabular}
\end{table}

\textbf{Time-MQA (Full FT)}: the same Time-MQA recipe (numeric series serialized to text via the BPE tokenizer, jointly trained on all six tasks) but with full-parameter fine-tuning instead of LoRA, isolating whether more trainable capacity rescues the text-serialization interface.

\textbf{ITFormer} \cite{wang2025itformer}: a QFormer-like \cite{li2023blip} architecture designed for temporal-textual question answering on the EngineMT-QA dataset. Since its original time series encoder is a pretrained PatchTST model that requires fixed-length multivariate input, which is incompatible with the flexible-length univariate time series in Time-MQA, we replace it with a frozen Time-MOE encoder \cite{shi2025time} and modify the cross-attention module to handle univariate time series.

\textbf{Frozen Time-MoE}: Replaces the trainable linear time-series encoder
with a pretrained, frozen Time-MoE encoder \cite{shi2025time}, while keeping
the same MLP projector and LoRA-tuned LLM; only the projector and LoRA
adapters are updated, under the cross-entropy loss alone. We freeze Time-MoE
following the setting of LLaVA~\cite{liu2023visual}, where the time-series
encoder is pretrained and kept frozen during QA fine-tuning, so that only the
cross-modal components adapt to the task. This isolates whether a large
pretrained time-series foundation model yields a stronger frozen
representation than a lightweight learnable encoder.

\subsection{Metrics}

For the forecasting and imputation tasks, we adopt three metrics. Throughout, let $N$ denote the number of samples and $\mathbb{1}[\cdot]$ the indicator function, which is $1$ when its condition holds and $0$ otherwise.

\noindent\textbf{(1) Format Compliance Rate (FCR)}, defined as the ratio of predictions whose output length exactly matches the requested length to the total number of predictions:
\begin{equation}
\mathrm{FCR}
=
\frac{1}{N}
\sum_{i=1}^{N}
\mathbb{1}
\!\left[
L(\hat{\mathbf{y}}_i)=L_i
\right],
\end{equation}
where $L_i$ is the required output length and $L(\hat{\mathbf{y}}_i)$ is the length of the generated output.

\noindent\textbf{(2) Own MSE}: the mean squared error computed over only those values a model actually predicts. As LLMs generate predictions as free-form text, they frequently emit fewer points than requested; for instance, producing $10$ values when $21$ are required. In such cases, Own MSE is evaluated solely on the predicted points (the first $10$):
\begin{equation}
\mathrm{Own\text{-}MSE}
=
\frac{1}{N_c}
\sum_{i=1}^{N_c}
\frac{1}{T_i}
\sum_{t=1}^{T_i}
(\hat y_{i,t}-y_{i,t})^2 ,
\end{equation}
where $N_c$ denotes the number of matched-length predictions and $T_i$ is the prediction length for sample $i$.

\noindent\textbf{(3) Shared MSE}, the mean squared error computed on the intersection of samples for which \emph{all} compared models produce correctly formatted outputs, enabling a head-to-head comparison on identical inputs:
\begin{equation}
\mathrm{Shared\text{-}MSE}
=
\frac{1}{N_s}
\sum_{i=1}^{N_s}
\frac{1}{T_i}
\sum_{t=1}^{T_i}
(\hat y_{i,t}-y_{i,t})^2 ,
\end{equation}
where $N_s$ denotes the number of samples for which all compared models produce format-compliant outputs.

For the remaining four tasks (classification, anomaly detection, true/false, and multiple choice), we use \textbf{Accuracy} as the evaluation metric, defined as
\begin{equation}
\mathrm{Accuracy}
=
\frac{1}{N}
\sum_{i=1}^{N}
\mathbb{1}
\!\left[
\hat c_i=c_i
\right],
\end{equation}
where $\hat c_i$ and $c_i$ are the predicted and ground-truth labels for sample $i$, respectively.

\subsection{Implementation Details}

Table~\ref{tab:impl_details} summarizes the key training configurations of our model. We build on a compact Qwen-3-0.6B backbone fine-tuned with LoRA, pairing it with a lightweight $1\!\rightarrow\!384$ trainable linear time-series encoder and an MLP projector, and train for $2{,}000$ steps with an effective batch size of $32$ on a single NVIDIA A100. The auxiliary one-directional supervised contrastive loss uses a memory bank of size $512$ and a loss weight of $\lambda = 0.1$.

\subsection{Main Results}

Table~\ref{tab:main_results} reports performance across all six Time-MQA tasks. Shared MSE is computed on the intersection of format-compliant predictions across all models except ITFormer, whose zero compliance rate would otherwise result in an empty shared subset. The shared subsets contain 79 forecasting samples and 148 imputation samples. CADE attains the best score on five of the ten reported metrics, 
forecasting FCR and own MSE, imputation own MSE, anomaly detection,
and MCQ, and remains competitive on the rest, whereas no baseline is
consistently strong across both the generative (forecasting, imputation) and
understanding (AD, classification, judgment, MCQ) task families.

\textbf{Query-compression discards the resolution generative tasks require.} The
QFormer-style ITFormer compresses each series into a fixed, small set of learned
query tokens, and this bottleneck is fatal on the generative tasks: it attains
$\mathrm{FCR}=0$ on both forecasting and imputation, never emitting a single
length-matched sequence. Its reported own-MSE values ($383{,}107$ for
forecasting, $4{,}284{,}782$ for imputation) are therefore not deployable
predictions, and it admits no shared subset at all. Because a fixed number of
query tokens cannot encode a variable-length series at per-timestep granularity,
the model loses exactly the positional structure needed to write back an output
aligned to the requested horizon, whereas our per-timestep linear encoding
preserves it.

\begin{table*}
\centering
\caption{Main results on the Time-MQA benchmark. Internal variants are included for completeness and analyzed in Section~\ref{sec:arch_abl}.}
\label{tab:main_results}
\scalebox{0.88}{
\begin{tabular}{l|ccc|ccc|cccc}
\toprule
\multirow{2}{*}{Method} & \multicolumn{3}{c|}{Forecasting} & \multicolumn{3}{c|}{Imputation} & \multirow{2}{*}{AD} & \multirow{2}{*}{Classification} & \multirow{2}{*}{True/False} & \multirow{2}{*}{MCQ} \\
& FCR $\uparrow$ & Own (MSE) $\downarrow$ & Shared (MSE) $\downarrow$ & FCR $\uparrow$ & Own (MSE) $\downarrow$ & Shared (MSE) $\downarrow$ & & & & \\
\midrule
ITFormer & 0 & 383,107 & --- & 0 & 4,284,782 & --- & 0.84 & 0.7525 & \textbf{0.7775} & 0.5013 \\
Frozen Time-MoE & 0.58 & 311,892 & 40,660 & \textbf{0.82} & 27,437 & 5,911 & 0.8475 & 0.77 & 0.7575 & 0.5113 \\
Frozen Random Linear & 0.554 & 315,402 & \textbf{29,917} & 0.7975 & 87,578 & 5,798 & 0.82 & \textbf{0.785} & 0.7550 & 0.5013 \\
Time-MQA & 0.46 & 440,626 & 30,757 & 0.65 & 2,399,043 & 5,318 & 0.5975 & 0.68 & 0.6775 & 0.471 \\
Time-MQA (Full FT) & 0.46 & 1,104,244 & 3,479,036 & 0.57 & 2,391,803 & \textbf{4,313} & 0.62 & 0.695 & 0.7 & 0.4332 \\
\midrule
CADE w/o SupCon & 0.596 & 312,649 & 36,852 & 0.777 & 34,532 & 6,004 & 0.835 & 0.7625 & 0.75 & 0.5189 \\
\textbf{CADE} & \textbf{0.598} & \textbf{296,897} & 32,268 & 0.785 & \textbf{25,210} & 5,999 & \textbf{0.8625} & 0.7675 & 0.7675 & \textbf{0.5315} \\
\bottomrule
\end{tabular}
}
\end{table*}

\textbf{On the shared-subset metric.} CADE does not attain the lowest
shared MSE (forecasting: $32{,}268$ vs.\ Frozen Random Linear's $29{,}917$;
imputation: $5{,}999$ vs.\ Time-MQA (Full FT)'s $4{,}313$). However, the shared subset is
restricted to rows that \emph{every} model formats correctly, and is therefore
bottlenecked by the lowest-FCR models and dominated by the easiest cases. Read
alongside the markedly higher format-compliance rates and lower own MSE of
CADE, these shared-subset errors indicate that our model
matches the baselines on easy inputs while additionally producing accurate,
correctly formatted predictions on the substantially larger and harder set of
inputs that the baselines fail to handle.

\subsection{Ablation Studies}

We organize our ablations into two groups. \emph{Architectural ablations}
(Section~\ref{sec:arch_abl}) isolate the contribution of each design choice in
our model, the time-series encoder and the auxiliary contrastive loss, by
degrading one component at a time while holding everything else fixed.
\emph{Hyperparameter ablations} (Section~\ref{sec:hp_abl}) then sweep the two
components introduced by the contrastive loss, the memory-bank size and the loss
weight $\lambda$.

\subsubsection{Architectural Ablations}
\label{sec:arch_abl}

To attribute CADE's gains to individual design choices, we introduce two
internal variants that each degrade a single component of the full model, and
analyze them against the external baselines using the numbers reported in
Table~\ref{tab:main_results}.

\textbf{Frozen Random Linear}: Replaces the trainable linear encoder with a randomly-initialized linear encoder that is held frozen throughout training, again updating only the projector and LoRA adapters under the cross-entropy loss. This baseline isolates the benefit contributed by learning the encoder, as opposed to the benefit that comes purely from the random projection itself.

\textbf{CADE w/o SupCon}: Our full architecture but supervised by the cross-entropy loss only. This isolates the contribution of the auxiliary contrastive loss in our final method.

\begin{table*}
\centering
\caption{Effect of memory-bank size ($\lambda = 0.1$). The first row is the
no-auxiliary-loss reference. Bold marks the best value among the swept settings
per column.}
\label{tab:abl_memory}
\scalebox{0.88}{
\begin{tabular}{l|cc|cc|cccc}
\toprule
\multirow{2}{*}{Memory size} & \multicolumn{2}{c|}{Forecasting} & \multicolumn{2}{c|}{Imputation} & \multirow{2}{*}{AD} & \multirow{2}{*}{Classification} & \multirow{2}{*}{True/False} & \multirow{2}{*}{MCQ} \\
& FCR $\uparrow$ & MSE $\downarrow$ & FCR $\uparrow$ & MSE $\downarrow$ & & & & \\
\midrule
None (no SupCon loss) & 0.596 & $312{,}649$ & 0.777 & $34{,}532$ & 0.835 & 0.7625 & 0.75 & 0.5189 \\
\midrule
32   & 0.52   & $290{,}927$ & 0.785 & $33{,}269$ & 0.8475 & \textbf{0.7675} & 0.7625 & 0.5264 \\
64   & 0.5171 & $314{,}575$ & 0.78  & $27{,}391$ & 0.8525 & 0.7625 & \textbf{0.77} & \textbf{0.5466} \\
128  & 0.546  & $286{,}400$ & 0.77  & $39{,}076$ & 0.8450 & 0.7575 & \textbf{0.77} & 0.5214 \\
256  & 0.58   & $278{,}346$ & 0.7675 & $29{,}254$ & 0.8525 & 0.76 & 0.76 & 0.5365 \\
\textbf{512}  & \textbf{0.598} & $296{,}897$ & 0.785 & \textbf{25{,}210} & \textbf{0.8625} & \textbf{0.7675} & 0.7675 & 0.5315 \\
1024 & 0.588  & $286{,}365$ & 0.785 & $30{,}576$ & 0.8525 & \textbf{0.7675} & 0.7625 & 0.5315 \\
2048 & 0.572  & \textbf{208{,}202} & \textbf{0.7875} & $32{,}849$ & 0.84 & 0.7625 & 0.7525 & 0.539 \\
\bottomrule
\end{tabular}
}
\end{table*}

\begin{table*}
\centering
\caption{Effect of contrastive loss weight $\lambda$ (memory size $= 512$). The
$\lambda = 0$ row is the no-auxiliary-loss reference. Bold marks the best value
among the active ($\lambda > 0$) settings per column.}
\label{tab:abl_lambda}
\scalebox{0.88}{
\begin{tabular}{l|cc|cc|cccc}
\toprule
\multirow{2}{*}{$\lambda$} & \multicolumn{2}{c|}{Forecasting} & \multicolumn{2}{c|}{Imputation} & \multirow{2}{*}{AD} & \multirow{2}{*}{Classification} & \multirow{2}{*}{True/False} & \multirow{2}{*}{MCQ} \\
& FCR $\uparrow$ & MSE $\downarrow$ & FCR $\uparrow$ & MSE $\downarrow$ & & & & \\
\midrule
0     & 0.596 & $312{,}649$ & 0.777 & $34{,}532$ & 0.835 & 0.7625 & 0.75 & 0.5189 \\
\midrule
0.01  & 0.591 & $310{,}394$ & \textbf{0.7875} & $33{,}929$ & 0.8575 & 0.76 & \textbf{0.7875} & 0.5416 \\
0.05  & 0.559 & $293{,}245$ & \textbf{0.7875} & $33{,}209$ & 0.84 & \textbf{0.7675} & 0.7675 & \textbf{0.5491} \\
\textbf{0.1}   & \textbf{0.598} & $296{,}897$ & 0.785 & \textbf{25{,}210} & \textbf{0.8625} & \textbf{0.7675} & 0.7675 & 0.5315 \\
0.25  & 0.591 & \textbf{282{,}121} & 0.7825 & $35{,}356$ & 0.8525 & 0.755 & 0.7675 & 0.5390 \\
0.5   & 0.572 & $313{,}433$ & \textbf{0.7875} & $29{,}219$ & 0.8475 & \textbf{0.7675} & 0.7425 & 0.5390 \\
\bottomrule
\end{tabular}
}
\end{table*}

\textbf{A continuous linear encoder substantially outperforms BPE tokenization.}
The sharpest contrast is between Time-MQA, which serializes the numeric series
into text and feeds it through the LLM's BPE tokenizer, and the encoder-based
variants, which inject the series as continuous embeddings. Replacing BPE with our trainable linear encoder improves FCR and own MSE on \emph{every} task: forecasting FCR rises
from $0.46$ to $0.596$ and its own MSE falls from $440{,}626$ to
$312{,}649$; imputation's own MSE drops from $2{,}399{,}043$ to $34{,}532$, nearly two orders of magnitude, and with the time series understanding tasks (anomaly detection, classification, judgment, MCQ) improving in the same direction (Table~\ref{tab:main_results}). Crucially, these gains already hold for ``CADE w/o SupCon,'' which differs from Time-MQA
\emph{only} in the continuous encoder, isolating the BPE bypass as the dominant
driver. The effect is in fact independent of \emph{learning} the encoder at all:
the Frozen Random Linear baseline, whose $1\!\rightarrow\!384$ projection is
randomly initialized and never trained, already surpasses BPE serialization on FCR and own MSE for every task, raising forecasting FCR to $0.554$ and cutting its own-MSE to
$315{,}402$, raising imputation FCR to $0.7975$ and collapsing its own-MSE from
$2{,}399{,}043$ to $87{,}578$, and lifting anomaly detection ($0.5975 \rightarrow
0.82$), classification ($0.68 \rightarrow 0.785$), judgment ($0.6775 \rightarrow
0.755$), and MCQ ($0.471 \rightarrow 0.5013$). Since a fixed random projection
carries no learned temporal information, this isolates the continuous-token
\emph{interface} itself, rather than any encoder capacity, as the source of the
improvement over BPE. The simultaneous collapse of Time-MQA's FCR and own MSE further suggests that tokenizing long numeric strings both
inflates sequence length and destroys the positional structure needed to emit
length-matched predictions. Full-parameter fine-tuning of the same BPE-serialized model (Time-MQA (Full FT)) yields mixed results and does not consistently improve performance (e.g.\ forecasting own-MSE worsens from $440{,}626$ to $1{,}104{,}244$ and MCQ drops from $0.471$ to $0.4332$, despite modest gains on several understanding metrics and imputation shared MSE), suggesting that increased trainable capacity alone does not resolve the limitations of the BPE.

\textbf{Simplicity is sufficient: a linear encoder rivals a pretrained
foundation encoder.} To isolate the effect of the encoder from that of the
auxiliary contrastive loss, we compare ``CADE w/o SupCon'' against the
frozen Time-MoE baseline, since neither uses the contrastive objective. Despite
its minimal capacity, a single $1\!\rightarrow\!384$ linear map, the trainable
encoder is broadly competitive with the frozen pretrained Time-MoE foundation
encoder: it is better on forecasting FCR ($0.596$ vs.\ $0.58$), forecasting
shared MSE ($36{,}852$ vs.\ $40{,}660$), and MCQ ($0.5189$ vs.\ $0.5113$),
essentially tied on forecasting own-MSE ($312{,}649$ vs.\ $311{,}892$),
classification ($0.7625$ vs.\ $0.77$), and true/false ($0.75$ vs.\ $0.7575$),
while trailing on anomaly detection ($0.835$ vs.\ $0.8475$) and imputation
own-MSE ($34{,}532$ vs.\ $27{,}437$).

\textbf{The one-Directional SupCon loss strengthens cross-modal alignment across tasks.}
Comparing ``CADE w/o SupCon'' with the full model isolates the auxiliary
one-directional supervised contrastive loss, which pulls projected time-series
embeddings toward frozen class-text anchors while leaving the text side fixed.
Although this signal is applied \emph{only} to classification samples, the
improvements are not confined to classification: anomaly detection rises from
$0.835$ to $0.8625$, judgment from $0.75$ to $0.7675$, MCQ from $0.5189$ to
$0.5315$, and imputation own MSE falls from $34{,}532$ to $25{,}210$ (a
$27\%$ reduction). This spillover indicates that anchoring time-series features
to the LLM's lexical embedding space regularizes the shared projection rather
than merely sharpening the anchored task, thereby strengthening the semantic
correspondence between time-series representations and the model's linguistic
reasoning across the board.

\subsubsection{Hyperparameter Ablations}
\label{sec:hp_abl}

We ablate the two hyperparameters introduced by our auxiliary contrastive loss: the memory size and the loss weight $\lambda$. Additionally, we omit Shared MSE in this section because the shared-subset MSE is inherently a cross-model metric. It is computed on the rows that all competing models format correctly, so its denominator is defined only with respect to a fixed set of distinct models. In an ablation study, the compared systems are variants of the same method rather than distinct models, making Shared MSE unsuitable for this purpose.

\textbf{Performance is robust across memory-bank sizes.} Table~\ref{tab:abl_memory}
shows the same stability with respect to the bank capacity: across the
$32$--$2048$ range, anomaly detection stays within $0.84$--$0.8625$ and
classification within $0.7575$--$0.7675$, with no setting degrading anomaly detection, judgment, or MCQ relative to the no-loss reference, and classification accuracy remaining nearly unchanged across settings. Individual
metrics peak at different sizes, forecasting own-MSE is lowest at $2048$ and MCQ
is highest at $64$, but these isolated extrema do not transfer to the other
metrics (e.g.\ at $2048$ both anomaly detection and judgment fall), and several
of the differences are at the scale of one or two test samples. We therefore
read the sweep as evidence of robustness rather than a sharp global optimum.

\begin{table*}
\centering
\caption{Comparison against a frontier general-purpose LLM (DeepSeek-V3.2) on the
Time-MQA benchmark.}
\label{tab:deepseek_compare}
\scalebox{0.88}{
\begin{tabular}{l|ccc|ccc|cccc}
\toprule
\multirow{2}{*}{Method} & \multicolumn{3}{c|}{Forecasting} & \multicolumn{3}{c|}{Imputation} & \multirow{2}{*}{AD} & \multirow{2}{*}{Classification} & \multirow{2}{*}{True/False} & \multirow{2}{*}{MCQ} \\
& FCR $\uparrow$ & Own (MSE) $\downarrow$ & Shared (MSE) $\downarrow$ & FCR $\uparrow$ & Own (MSE) $\downarrow$ & Shared (MSE) $\downarrow$ & & & & \\
\midrule
DeepSeek-V3.2 & \textbf{0.984} & \textbf{270,979} & 85,688 & \textbf{0.91} & 87,334 & 120,663 & 0.5975 & 0.6875 & 0.7475 & \textbf{0.585} \\
\textbf{CADE} & 0.598 & 296,897 & \textbf{82,458} & 0.785 & \textbf{25,210} & \textbf{31,200} & \textbf{0.8625} & \textbf{0.7675} & \textbf{0.7675} & 0.5315 \\
\bottomrule
\end{tabular}
}
\end{table*}

\textbf{Turning the loss on helps; the method is not sensitive to its exact
weight.} The clearest signal in Table~\ref{tab:abl_lambda} is the gap between
$\lambda = 0$ and \emph{any} positive weight on the understanding tasks: every
active setting raises anomaly detection above the $0.835$ obtained without the
loss (to $0.84$--$0.8625$), and likewise improves MCQ ($0.5189 \rightarrow
0.5315$--$0.5491$) and, for most settings, judgment. Because the loss is applied
only to classification samples, these consistent cross-task gains indicate that
anchoring time-series embeddings to the lexical space regularizes the shared
projection rather than over-fitting the anchored task. Importantly, performance
remains stable a $50\times$ range in $\lambda$ with no collapse,
showing the contribution of the loss is robust rather than the artifact of a
single fortunate weight.

\textbf{$\lambda = 0.1$ and memory size $512$ give the best overall balance.}
The two sweeps agree on a single joint configuration. Anomaly detection, the
most semantically structured understanding task, and the one the contrastive
alignment should most directly benefit, reaches its peak of $0.8625$ at exactly
$\lambda = 0.1$ in Table~\ref{tab:abl_lambda} and at exactly memory size $512$ in
Table~\ref{tab:abl_memory}; no other setting attains this value in either sweep.
This same configuration also yields the best forecasting compliance (FCR $0.598$), the lowest imputation own-MSE among all swept settings ($25{,}210$, a
clear margin over the next-best $\sim\!27{,}000$--$30{,}000$), and ties for the
best classification accuracy ($0.7675$), while remaining competitive on the remaining metrics.

\subsection{Comparison with a Frontier General-Purpose LLM}

To situate our compact, encoder-based model against a strong text-only baseline,
we additionally evaluate DeepSeek-V3.2, a frontier general-purpose LLM that
ingests the numeric series via text serialization and is prompted zero-shot
(no time-series encoder, no task-specific tuning). Because the two systems have
very different format-compliance rates, the shared subset here is recomputed
\emph{pairwise} over the intersection of the two models' format-compliant
predictions ($221$ samples for forecasting, $288$ for imputation), and therefore
differs from the all-model shared subset in Table~\ref{tab:main_results}.

\textbf{Scale buys format compliance, not numeric accuracy.} DeepSeek-V3.2
attains near-perfect format compliance (FCR $0.984$ on forecasting, $0.91$ on
imputation) against our $0.598$ and $0.785$, confirming that a large
instruction-tuned model is far more reliable at emitting length-matched
sequences. Yet once the comparison is restricted to inputs both models format
correctly, this scale advantage does not translate into accuracy: on the shared
subset our $0.6$B model already edges out DeepSeek on forecasting ($82{,}458$
vs.\ $85{,}688$) and is nearly $4\times$ lower on imputation ($31{,}200$ vs.\
$120{,}663$). The same holds for imputation own-MSE ($25{,}210$ vs.\ $87{,}334$),
while the two are roughly comparable on forecasting own-MSE
($296{,}897$ vs.\ $270{,}979$, the latter measured over DeepSeek's much larger
compliant set). In other words, presenting the series as continuous tokens yields
more accurate numeric predictions than text serialization even against a model
orders of magnitude larger.

\textbf{Discriminative time-series understanding favors the continuous encoder.}
The gap widens sharply on the tasks that require reading temporal structure
rather than copying numbers. CADE outperforms DeepSeek-V3.2 on anomaly
detection by a wide margin ($0.8625$ vs.\ $0.5975$), and also on classification
($0.7675$ vs.\ $0.6875$) and true/false judgment ($0.7675$ vs.\ $0.7475$). A
frontier text model that only sees serialized digits struggles to localize
anomalies, whereas the trainable encoder exposes the underlying shape of the
series. This indicates that the principal bottleneck for general LLMs on time
series is \emph{representation}, not reasoning capacity.

\textbf{MCQ is the exception.} The one task where DeepSeek leads is multiple
choice ($0.585$ vs.\ $0.5315$), which leans most heavily on broad reasoning and
world knowledge rather than on fine-grained reading of the series. Here the
limited capacity of the $0.6$B backbone is visible, and the larger model's
general competence dominates.

\subsection{Limitations and Future Work}

\textbf{Format compliance remains a bottleneck.} Although the continuous encoder
markedly improves FCR over small-model BPE serialization
(Table~\ref{tab:main_results}), our compliance still trails a frontier model by a
large margin (forecasting FCR $0.598$ vs.\ $0.984$; imputation $0.785$ vs.\
$0.91$). Because forecasting and imputation outputs are only usable when their
length matches the request, this gap directly caps the fraction of inputs on
which our model produces a deployable prediction, even though its accuracy on the
compliant subset is competitive or better.

\textbf{Reasoning-heavy MCQ lags.} CADE underperforms on MCQ, the most
reasoning-intensive task, which suggests that neither the small backbone nor the
current alignment signal is sufficient for multi-option discrimination that
combines temporal evidence with broader inference.

\textbf{Alignment is too implicit.} Our auxiliary one-directional supervised
contrastive loss is applied \emph{only} to classification samples and anchors
time-series embeddings to a fixed set of class-text vectors. While this provides
a useful regularizer for all tasks, the model is
never explicitly taught what fundamental temporal concepts such as trend,
periodicity, level shifts, or anomalies \emph{mean} in language. We hypothesize
that a more principled, two-stage curriculum would close both the FCR and MCQ
gaps. Concretely, rather than relying on a contrastive term during QA tuning, the
model should first be pretrained on large-scale time-series\,--\,text paired data
that describe these basic concepts, learning to map temporal structure onto its
lexical space, and only then be fine-tuned on the downstream QA tasks. Such
explicit concept alignment, followed by task adaptation, would give the model a
genuine semantic grounding of time-series structure instead of a narrow auxiliary
signal, which we leave to future work.
\section{Conclusion}
In this work, we presented CADE (Contrastive Alignment with Direct Embedding), a framework for multi-task time-series question answering that maps each timestep directly into the LLM's embedding space through a linear encoder and MLP projector, and aligns these embeddings with frozen class-text anchors via a one-directional supervised contrastive loss. Experiments on the Time-MQA benchmark show that CADE consistently outperforms BPE serialization and remains competitive with or better than both a pretrained foundation encoder and a frontier general-purpose LLM across the six tasks; our ablation studies further indicate that the continuous-token interface is the dominant driver of these gains, while the auxiliary contrastive loss contributes consistent cross-task improvements, most notably on the understanding tasks. Since format compliance and reasoning-heavy questions remain the main limitations, future work will explore more explicit semantic grounding of temporal structure prior to downstream task adaptation.

\section*{Acknowledgment}
This research was funded (partially or fully) by the Australian Government through the Australian Research Council. Dr Hung Le is the recipient of an Australian Research Council Discovery Early Career Researcher Award (project number DE250100355) funded by the Australian Government.

\bibliographystyle{IEEEtran}
\bibliography{papers}

\end{document}